\documentclass[letterpaper, 10 pt, conference]{ieeeconf}  

\IEEEoverridecommandlockouts                              
\usepackage[utf8]{inputenc}
\usepackage[T1]{fontenc}
\usepackage{multirow}
\usepackage{graphicx}
\usepackage{svg}
\usepackage{amsmath}
\DeclareMathOperator*{\argmin}{arg\,min}
\DeclareMathOperator*{\argmax}{arg\,max}
\usepackage{amsfonts,amssymb}
\usepackage{xcolor}
\usepackage[ruled,linesnumbered]{algorithm2e}

\title{\LARGE \bf
3D Perception based Imitation Learning under Limited Demonstration for Laparoscope Control in Robotic Surgery 
}

\author{Bin Li$^{1}$, Ruofeng Wei$^{2}$, Jiaqi Xu$^{3}$, Bo Lu$^{1}$, Chi Hang Yee$^{4}$, Chi Fai Ng$^{4}$, \\
Pheng-Ann Heng$^{3}$, \textit{Senior Member, IEEE}, Qi Dou$^{3}$, \textit{Member, IEEE}, and Yun-Hui Liu$^{1}, \textit{Fellow, IEEE}$
\thanks{*This work is supported in part of the HK RGC under T42-409/18-R and 14202918,  in part by Shenzhen Portion of Shenzhen-Hong Kong Science and Technology Innovation Cooperation Zone under HZQB-KCZYB-20200089, in part by the Multi-Scale Medical Robotics Centre, InnoHk, and in part by the VC Fund 4930745 of the CUHK T Stone Robotics Institute.}
\thanks{$^{1}$B. Li, B. Lu, and Y. H. Liu are with the T stone Robotics Institute, The Department of Mechanical and Automation Engineering, The Chinese University of Hong Kong.}%
\thanks{$^{2}$R. Wei is with the Department of Biomedical Engineering, City University of Hong Kong, Hong Kong.}%
\thanks{$^{3}$J. Xu, P.A. Heng, Q. Dou are with the Department of Computer Science and Engineering, The Chinese University of Hong Kong.}%
\thanks{$^{4}$C.H. Yee and C.F. Ng are with the SH Ho Urology Centre, Department of Surgery, The Chinese University of Hong Kong.}%
\thanks{Corresponding author: Bo Lu (bolu@cuhk.edu.hk)}
}

\begin{document}

\maketitle
\thispagestyle{empty}
\pagestyle{empty}

\begin{abstract}


Automatic laparoscope motion control is fundamentally important for surgeons to efficiently perform operations.
However, its traditional control methods based on tool tracking without considering information hidden in surgical scenes are not intelligent enough, while the latest supervised imitation learning (IL)-based methods require expensive sensor data and suffer from distribution mismatch issues caused by limited demonstrations.
In this paper, we propose a novel Imitation Learning framework for Laparoscope Control (ILLC) with reinforcement learning (RL), which can efficiently learn the control policy from limited surgical video clips.
Specially, we first extract surgical laparoscope trajectories from unlabeled videos as the demonstrations and reconstruct the corresponding surgical scenes.
To fully learn from limited motion trajectory demonstrations, we propose Shape Preserving Trajectory Augmentation (SPTA) to augment these data, and build a simulation environment that supports parallel RGB-D rendering to reinforce the RL policy for interacting with the environment efficiently.
With adversarial training for IL, we obtain the laparoscope control policy based on the generated rollouts and surgical demonstrations. 
Extensive experiments are conducted in unseen reconstructed surgical scenes, and our method outperforms the previous IL methods, which proves the feasibility of our unified learning-based framework for laparoscope control.


\end{abstract}

\section{INTRODUCTION} 
Robot-assisted minimally invasive surgery (RMIS) nowadays is widely used in a variety of laparoscopic procedures, which can reduce the patient's trauma and postoperative hospitalization~\cite{vitiello2012introduction}.
A high-quality field of view (FoV) from the laparoscope is crucial for providing surgeons with suitable real-time visual feedback throughout the duration of a laparoscopic surgery.
To help alleviate the burden of surgeons on periodically adjusting the laparoscope FoV, thus assisting them to improve surgical efficiency and performance, 
automatic laparoscope control has recently become popular for developing surgical automation techniques.
However, this task has two main challenges.
First, the laparoscope needs to be accurately controlled within a limited 3D space in body.
The second is intelligent perception of 3D surgical scenes for deploying generalizable laparoscope control policies under complex surgical scenarios~\cite{pandya2014review}.

\begin{figure}[tp]
  \centering
  \includegraphics[width = 1.0\hsize]{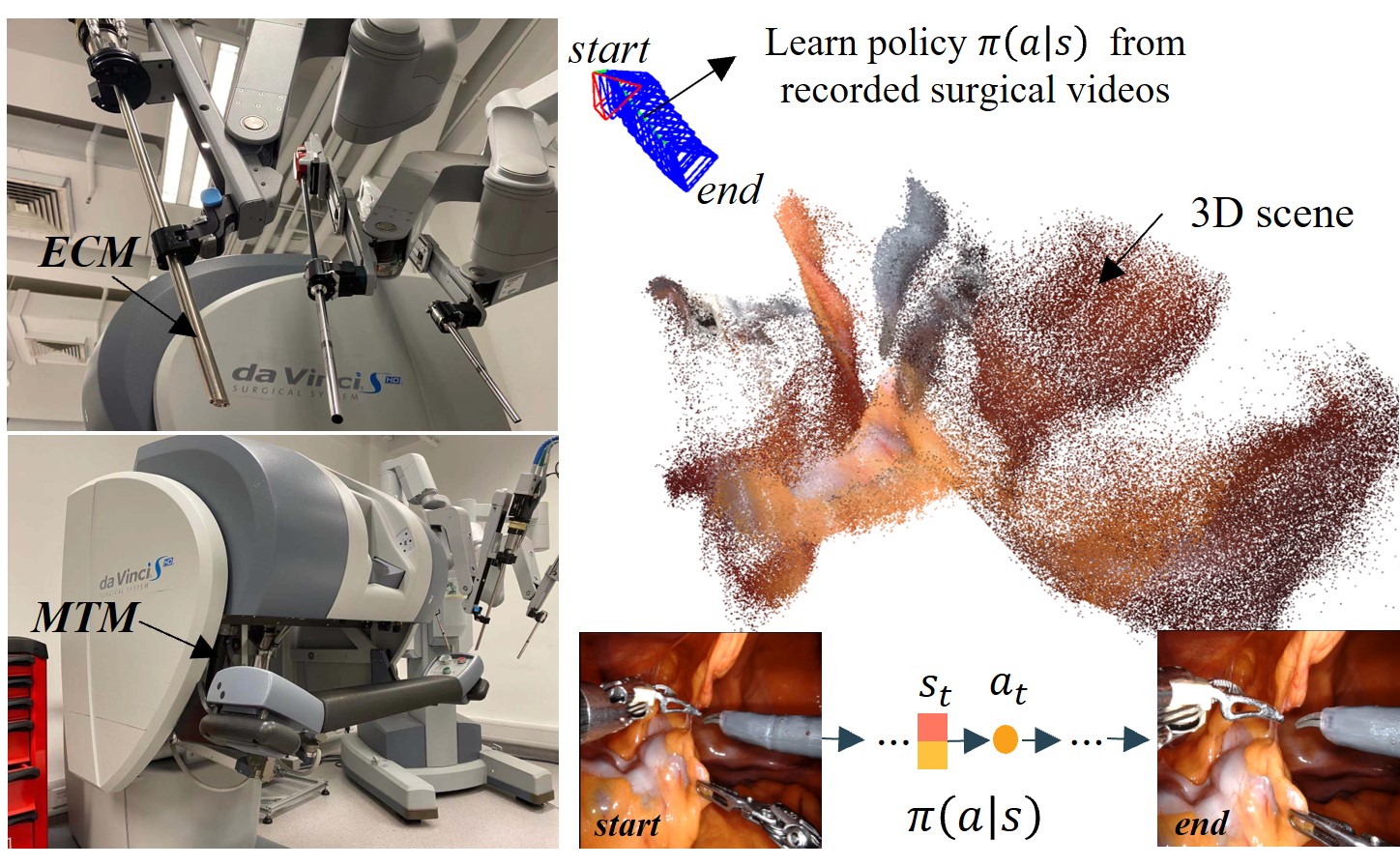}
  \caption{Surgeons operate the laparoscope (ECM) by MTM to obtain the  best FoV.
  Based on these limited demonstrations, we propose an ILLC to learn how to adjust FoV from raw recorded surgical videos.
  }
  \label{fig:overview of process}
\vspace{-0.5cm}
\end{figure}

To address these challenges, previous efforts concentrate on the rule-based laparoscope control schemes with diverse perception forms, such as tool tracking~\cite{osa2010framework,yang2019adaptive,li2021data}, eye gaze~\cite{fujii2018gaze, gras2017implicit}, etc.
For example, Osa~\textit{et~al.}~\cite{osa2010framework} proposed a visual servoing method that controlled the laparoscope to keep the surgical tool at the center point of FoV.
Li~\textit{et~al.}~\cite{li2021data} further extended the center point into a domain knowledge based area derived from procedural understanding of the surgical videos.
However, these hand-crafted rules have a limited level of intelligence and can hardly be applied to complex surgical environments.
Another solution is learning-based methods, which attempt to inject data-driven intelligence into the laparoscope control. 
Heuristics model~\cite{bihlmaier2014automated} and Gaussian mixture models (GMM)-based method~\cite{rivas2019transferring} are proposed to explore the connection between laparoscope motion and tool positions.
Nevertheless, such methods require expensive extra sensory data as the state information (e.g., recording poses of tools and laparoscope) for behavior learning, which is typically unavailable in recorded surgical videos in clinical routine.


Recent advances in imitation learning (IL) based on visual states modelled by convolutional neural network (CNN) have shown promise to imitate expert behaviors from raw image observations. Some impressive success based on such image features has been demonstrated on general robot tasks, such as autonomous driving~\cite{ly2020learning,xu2017end}, aerial filming~\cite{huang2019learning,huang2019one}, virtual cinematography~\cite{jiang2020example}.
However, learning automatic laparoscope view adjustment skills from surgical videos with raw RGB-D images as the observation is largely unexplored.
Moreover, the limited expert demonstration data extracted from the recorded videos poses challenges to learned models' generalizability to unseen scenarios.
Although some online IL methods such as DAgger~\cite{ross2011reduction} can alleviate this problem, it still requires expensive labeling with the surgeon's efforts, which is inefficient.
Very recently, another solution is to combine IL with reinforcement learning (RL) to achieve better generalizability through self-exploration, where the reward or policy can be recovered from expert demonstrations.
Chi~\textit{et~al.} show a GAIL-based approach for catheterization~\cite{chi2020collaborative}, and we also show that the bimanual peg transfer task can be learned with demonstrations~\cite{xu2021surrol}.
%
%
Nonetheless, these vanilla IL methods may still result in policies that suffer from low performance and generalizability with image observations.
Therefore, it is critical to design an efficient IL method using limited surgical expert data to learn the laparoscope control in realistic surgical environments.

In this paper, we propose a novel imitation learning framework for laparoscope control (ILLC), which to the best of our knowledge is the first to use IL with RL to effectively learn the laparoscope motion skills from limited real-world surgical videos, as shown in Fig.~\ref{fig:overview of process}.
Specifically, we first reconstruct 3D surgical scenes from surgical videos as the interactive RL environment and extract laparoscope motion trajectories as the expert demonstration data.
To efficiently learn from the limited demonstration, we propose a shape-preserving trajectory augmentation (SPTA) to augment the expert data and use a prior policy to bootstrap the learning process.
Afterward, the policy is reinforced using the adversarial learning approach through augmented demonstrations and continuous interactions in our simulated realistic surgical environment.
Our main contributions are three folds:

\begin{itemize}
\item We propose a novel ILLC framework with RL to achieve intelligent autonomous laparoscope control by learning the joint information between the laparoscope motion and surgical scenes from raw surgical videos.
\item We introduce a shape-preserving trajectory augmentation (SPTA) scheme and prior policy to overcome the limited demonstration problem in surgical field and bootstrap the training process to improve the performance of autonomous laparoscope control.
\item We conduct validation experiments on controlling the laparoscope under unseen realistic surgical scenes to show the feasibility of our proposed framework and its superiority compared with previous IL baselines.
\end{itemize}


\section{Problem formulation}
We aim at building a learner that can imitate surgeons' behavior for laparoscope view adjustment based on the observed surgical scene.
Formally, we model the laparoscope control problem as a Markov Decision Process (MDP), represented by $(\mathcal{S}, \mathcal{A}, \mathcal{R}, \mathcal{P}, p(s_0), \gamma)$, with state space $\mathcal{S}$, action space $\mathcal{A}$, reward function $\mathcal{R}$, transition probability $\mathcal{P}$, initial state distribution $p(s_0)$, and discount factor $\gamma$.
The objective of the learner is to find a policy $\pi_{\phi^*}(a|s)$ that can generate an action $a \in \mathcal{A}$ given the observed state $s \in \mathcal{S}$, and maximize the discounted sum of the rewards $r_{\theta}(s,a)$:
\begin{equation}
\max _{\pi_{\phi}} \mathbb{E}_{\pi_{\phi}}\left[\sum_{t=1}^{T} \gamma^{t} r_{\theta}\left(s_{t}, a_{t}\right)\right]
\end{equation}
where the parameter $\theta$ is recovered from expert demonstration trajectories $\mathcal{D}=\left\{\tau_{1}, \ldots, \tau_{N}\right\}$:
$\max _{\theta} E_{\tau \sim \mathcal{D}}\left[\log p_{\theta}(\tau)\right]$, $p_{\theta}(\tau) \propto p\left(s_{0}\right) \prod_{t=0}^{T} p\left(s_{t+1}|s_{t}, a_{t}\right) e^{\gamma^{t} r_{\theta}\left(s_{t}, a_{t}\right)}$.
%


In this paper, we propose an adversarial imitation learning based approach~\cite{fu2018learning} to learn the laparoscope control policy.
Specifically, we use a state-of-art RL method to train the learning agents, which collect rollouts within the environment by observing surgical scenes and generating actions.
Simultaneously, we train a discriminator to classify whether the trajectory is from expert demonstrations or agent rollouts.
In this way, the discriminator can recover the reward function $r_\theta$ from two different collected data sources, maximizing the reward for RL agents that imitate surgeons' behaviors.
Therefore, We obtain the IL policy by optimizing the RL agents and the discriminator alternatively until convergence, which is able to automate the laparoscope control and mimic the surgeon at the same time.

\begin{figure*}[thpb]
  \centering
  \includegraphics[width = 1.0\hsize]{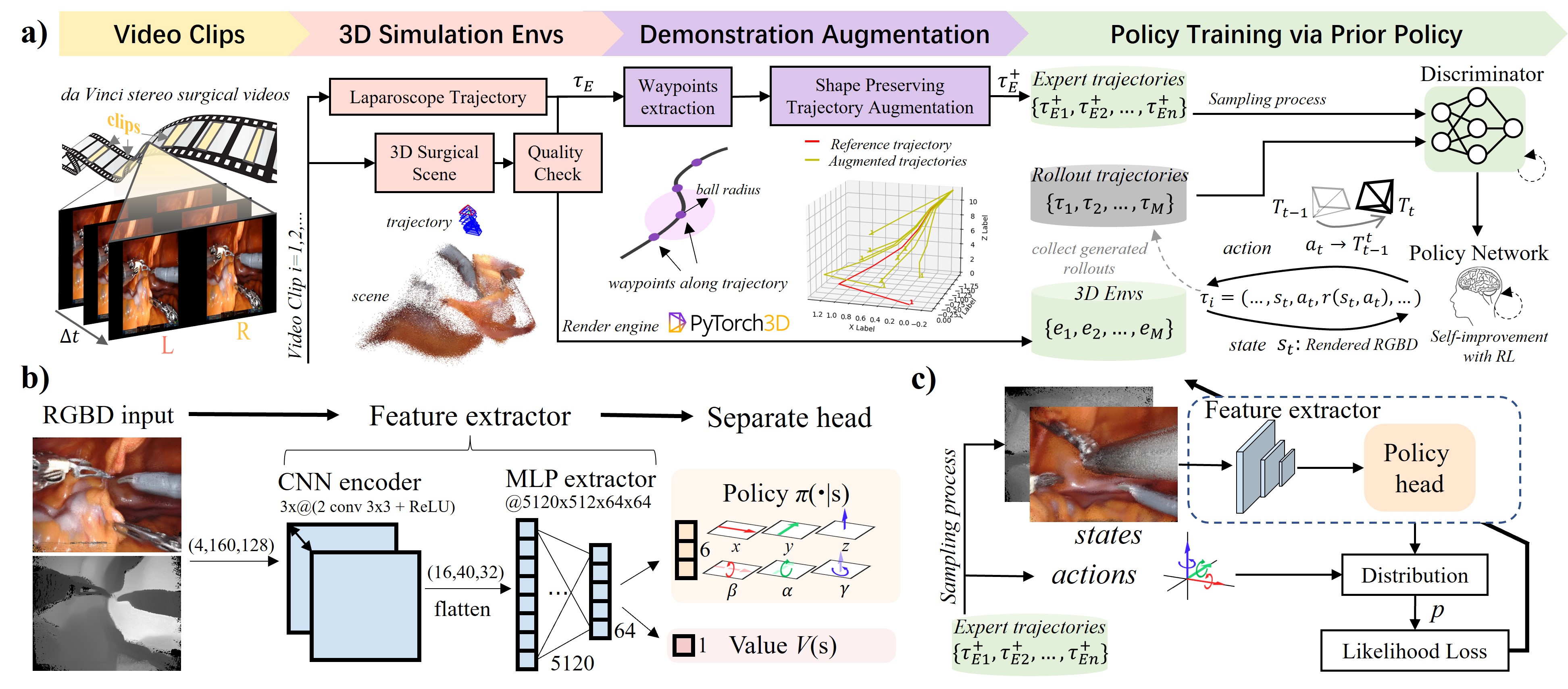}
  \caption{
  \textbf{The proposed ILLC framework for laparoscope control.}
  (a) Our ILLC framework comprises 3D scene reconstruction, expert data augmentation, and policy training.
  (b) The RL policy network consists of a shared CNN encoder, MLP feature extractor, and two separate policy and value heads.
  (c) Prior policy generated from expert demonstrations with likelihood loss.
  }
  \label{fig:overall framework}
\vspace{-0.5cm}
\end{figure*}

\section{METHODS}
This section describes our proposed ILLC framework for intelligent autonomous laparoscope control in detail, including 3D simulation environment construction, expert demonstration augmentation, and policy learning strategy, as shown in Fig.~\ref{fig:overall framework}.

\subsection{3D perception-based surgical scene rendering}
%
The first step is to build realistic surgical scenes based on the collected videos as the environment for RL agents to control laparoscope and interact within.
Meanwhile, the expert trajectories are estimated from their corresponding videos as demonstrations.
Specifically, the surgical environment is first reconstructed as the 3D object representation.
Afterward, the scene image viewed by the laparoscope is rendered via a 3D projection with the knowledge of the camera pose.

In the scene reconstruction, as the laparoscope and surgical tools cannot be manipulated simultaneously in the da Vinci surgical system~\cite{dimaio2011vinci}, we assume the surgical environment remains static when the laparoscope moves, which allows us to reconstruct the 3D scene based on the rigid body assumption.
For each frame, we first estimate the dense depth information based on the hierarchical deep stereo matching (HSM)~\cite{yang2019hierarchical} and calculate the camera pose using the iterative closest point (ICP), with the stereo calibration parameters~\cite{zhang2000flexible}.
Finally, the 3D reconstructed scene for each video clip is obtained by jointly merging the object from all frames.
You can refer to our previous work \cite{wei2021stereo} for more details.
The laparoscope motion clips are extracted from our HKPWH dataset~\cite{long2021dssr} containing real-world stereo prostatectomy videos performed by experienced surgeons.

\begin{figure}[tp]
    \centering
    \includegraphics[width = 1.0\hsize]{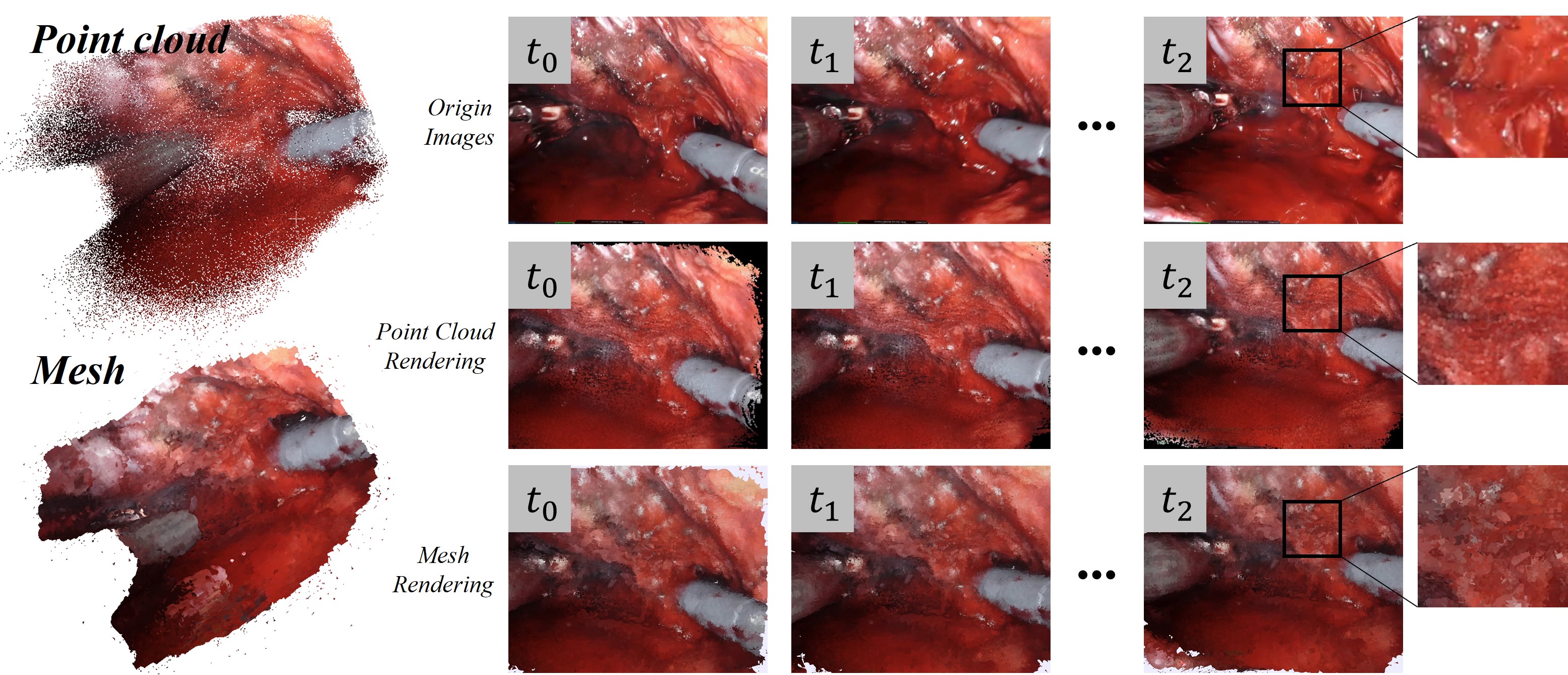}
    \caption{
    \textbf{Reconstructed surgical scene.}
    Left: Screenshots of the reconstructed surgical scenes in the point cloud modality (up) and the mesh modality (bottom).
    Right: The sequences are origin video image (up), rendered image at estimated camera positions in point cloud scene (middle) and mesh scene (bottom).
    }
    \label{fig:render}
    \vspace{-0.5cm}
\end{figure}

Based on the reconstructed scene, we build our ECM Simulation environment \textit{ECMSim} for laparoscope control, which can render realistic RGB-D images when the laparoscope moves within the environment at any position, with \textit{PyTorch3D}~\cite{ravi2020pytorch3d} as our rendering engine.
The dynamic physical interaction between the laparoscope and the tissue is not considered in this work.
To maximize the rendering realism, we compared two commonly-used 3D object representations, i.e., origin point cloud and its corresponding polygonal mesh, as illustrated in Fig.~\ref{fig:render}.

Although the polygonal mesh is theoretically superior to the point cloud, it needs to be generated from the high-quality point cloud.
Due to the short duration of the video clips and subtle estimation errors of the depth maps, the quality of the reconstructed point cloud is enough for rendering but suboptimal for high-quality mesh generation.
As in Fig.~\ref{fig:render}, the rendered images from the point cloud is almost the same with the original video images, which shows the accurate estimated trajectory and reconstructed environment scene.
Moreover, we can observe that the visualization performance from point cloud is superior than the rendered images with smearing effects from the generated mesh. 
Therefore, we adopt the reconstructed point cloud as the 3D object representation of the scene for rendering.

\subsection{Demonstration augmentation with SPTA} 
The expert state-action trajectories $\tau_E$ extracted during 3D scene reconstruction is of limited number, which may hinder the generalization of the learned model.
To solve this problem, we propose a shape-preserving trajectory augmentation (SPTA) method to augment the collected expert data and build an enriched expert demonstration database $\tau_E^+$.

\textbf{Raw expert trajectory preprocessing:}
For raw expert trajectories extracted during the 3D scene reconstruction, we observe that there exist local jittering fluctuations and inconsistent speeds.
Thus, we use the first-order Savitzky-Golay filter to smooth the raw expert trajectories, both in position $(x,y,z)$ and orientation $(\alpha,\beta,\gamma)$. To acquire consistent speed during motion, 
the smoothed trajectory of $m$ points $P_{i\in\{1,...,m\}}$ is then generated into $n$ equally-distributed waypoints $W_{j\in\{1,...,n\}}$ at a fixed separation distance $d_{fixed}$, based on the linear interpolation as follows:
\begin{equation}
\label{equ: traj_process}
    \begin{split}
    W_{j}^{\left(\cdot\right)} &=  W_{j-1}^{\left(\cdot\right)} + \frac{d_{fixed}}{\left\|P_{i^*}^{\left(\cdot\right)} - W_{j-1}^{\left(\cdot\right)}\right\|_2} (P_{i^*}^{\left(\cdot\right)} - W_{j-1}^{\left(\cdot\right)}), \\
    i^{*} &= \argmin _{i \in \{1,...,n\}}  (\left\|P_{i}^{\left(\cdot\right)} - W_{j-1}^{\left(\cdot\right)}\right\|_2 > d_{fixed}).
    \end{split}
\end{equation}
where the initial waypoint $W_0^{\left(\cdot\right)}$=$P_0^{\left(\cdot\right)}$, and $^{\left(\cdot\right)}$ can be any of $\{x,y,z,\alpha,\beta,\gamma\}$.
%
%
Then we calculate the $j^{th}$ waypoint's camera pose relative to the initial camera pose~$T_0$, represented by a transformation matrix $T_0^j$ with rotation matrix $R(\alpha,\beta,\gamma)$ and translation vector $t(x,y,z)$.
Its observed state $s_j$ as an RGB-D image is rendered by our \textit{ECMSim}.
The corresponding action $a_{j}$ is calculated as: $a_{j}\rightarrow T_{j}^{j+1}=inv(T_0^j) \cdot T_0^{j+1}$, and the action $a$ is defined as the transformation $(x,y,z,\alpha,\beta,\gamma)$ relative to the current camera pose.
%

\textbf{Shape Preserving  Trajectory  Augmentation (SPTA):}
%
To augment surgeons' demonstrations and mimic the expert behavior, we consider two criteria:

\noindent \textbf{Criterion 1:} The augmented laparoscope motion trajectory should maintain a similar movement pattern to the expert trajectory for each surgical scene.

\noindent \textbf{Criterion 2:} The laparoscope needs to stop near the endpoint in the expert trajectory to obtain a desirable view similar to the expert for any sampled start positions.

To enable this, we first generate a 3D cubic workspace $WS$ from the trajectory $\tau=W_{j\in\{1,...,end\}}$:
%
%
$WS = Cube\{W_{1:k}\}$, where $k$ is randomly chosen and slightly larger than 1, so the generated $WS$ is near the starting point $W_1$.
Then the starting point $S$ is uniformly sampled within $WS$ for diversity.
Afterward, we generate an augmented trajectory from $S$ to $W_{end}$ of a similar movement pattern with the original expert demonstration.

To achieve a balance between the local and global similarity, we acquire the waypoint point $W_{j^{*}}$ closest to sampled start point $S$ and use the trajectory segment from this point to the trajectory endpoint $W_{j^{*}:end}$ as the reference for augmentation.
Therefore, the influence of unused motion pattern in the initial part $W_{1:j^{*}}$ far from the start point can be avoided, which is similar to the idea of reward-to-go~\cite{tamar2016learning}.
Next, we treat it as a non-linear fitting problem that use the exponential decay to gradually reduce the initial distance $e_0$ between $S$ and $W_{j^{*}}$ as follows:
\begin{equation}
    \label{equ: traj_start}
    dist = \left\|S - W_{j^{*}}\right\|_2,\ j^{*} = \argmin_{j \in \{1,...,end\}} (\left\|S - W_{j}\right\|_2)
\end{equation}
\begin{equation}
    \label{equ: traj_aug}
    \begin{split}
    %
    %
    W'_{j'} &= W_{j^{*}+j'} + (1-\epsilon) \cdot \Delta e_{j'},\\
    \Delta e_{j'} &= k_{1} \cdot e^{\gamma \cdot j'} + k_{2},\ j'=0,1,...,end-j^{*}.
    \end{split}
\end{equation}
where the coefficients $k_{1}$, $k_{2}$ are solved with two boundary conditions: $e_{0} = dist$ and $e_{end} = \epsilon$, $\epsilon$ is a random small tolerance; $\gamma$ is the hyper-parameter which tunes the approaching speed, and $W'_{j'}$ is the augmented waypoint.

\subsection{Policy learning via prior policy}  
Inspired by~\cite{fu2018learning}, we use adversarial learning approach for policy learning with RGB-D image input in the ILLC framework.
The RL laparoscope agent policy $\pi_{\phi}\left(\cdot\right)$ parameterized by $\phi$ is modeled as the generator and is trained against the discriminator adversarially.
The discriminator $D_\theta$ is trained to classify surgeon demonstration trajectories $(s_{t}, a_{t}) \sim \tau_{E}^{+}$ and agent rollouts $(s_{t}^{\prime}, a_{t}^{\prime}) \sim \tau$, with the cross-entropy loss:
\begin{equation}
\label{equ: dism_training}
\begin{split}
L_{\theta}({D}) = -\ \mathbb{E}_{\tau_{E}^{+}} \left[\log \left( D_{\theta} \left(s_{t}, a_{t} \right) \right) \right] \\
-\ \mathbb{E}_{\tau} \left[ \log \left(1-D_{\theta} \left(s_{t}^{\prime}, a_{t}^{\prime} \right) \right) \right]
\end{split}
\end{equation}

With the following form for discriminator: 
$D_\theta(s,a)=\frac{\pi_{\phi}(a | s)}{\exp \{r_{\theta}(s, a)\} + \pi_{\phi}(a | s)} = f(\log \pi_{\phi}(a | s) - r_{\theta}(s,a))$,
where $f(\cdot)$ is the softmax function, $r_{\theta}(\cdot)$ is the reward function to be recovered.
Meanwhile, the policy $\pi_\phi$ is trained to minimize the entropy-regularized discriminative reward: $\hat{r}(s,a)=\log \pi_{\phi}(a|s)-r_{\theta}(s,a)$.
The entropy-regularized RL objective is obtained by summing $\hat{r}(s_t,a_t)$ over entire trajectories:
\begin{equation}
    \label{equ:ppo agent}
    \resizebox{0.9\hsize}{!}{$
    \mathbb{E}_{\pi_\phi}\left[\sum \hat{r} \left(s_{t}, a_{t}\right) \right] = \mathbb{E}_{\pi_\phi}\left[\sum \log \pi_\phi \left(a_{t}|s_{t} \right) - r_{\theta}\left(s_{t}, a_{t}\right) \right]$}
\end{equation}
And the global optima of the discriminator objective is achieved when $\pi_\theta=\pi_{E}$, as the learned policy $\pi_\theta$ converges to the expert policy $\pi_{E}$.

Considering that the current view alone can determine the next action in the laparoscope control, we assume that the ground truth reward is only related to the current state (RGB-D image), which meets the basis of the algorithm~\cite{fu2018learning}.
The stochastic Gaussian policy is used for continuous laparoscope actions,
$\pi_{\phi}(a|s)=\mu_{\phi}(s)+\sigma_{\phi}(s)\odot\epsilon$,
where mean and variance are generated by deterministic functions $\mu_{\phi}$, $\sigma_{\phi}$, $\epsilon\sim\mathcal{N}(0,1)$, and $\odot$ denotes the element-wise product.

During training, all training scenes are loaded into multiple threads simultaneously to accelerate the data collection, and each scene acts as an independent scene-agent environment.
We collect the agents' rollout trajectories from all training environments and combine them with expert demonstrations to optimize the discriminator $D_\theta$.
The accordingly updated rewards with $r_\theta$ in the previous rollouts are then used to update the agent policy $\pi_\phi$ using Eq.~\ref{equ:ppo agent}.

\textbf{Bootstrapping with prior policy:}
Generally, the adversarial training will gradually optimize the agent policy and the discriminator.
However, as laparoscope control requires high precision, if the policy starts from random, a large amount of valueless rollout in the early stage will make policy and discriminator get stuck with low performance.
To bootstrap the training process and sidestep the exploration challenge, we utilize the augmented expert demonstration $\pi_E^+$ to pre-train the policy and get a stable prior policy $\pi_{\phi_{init}}$, inspired by~\cite{he2019rethinking}, followed with the adversarial training to further fine-tune the parameters.
Formally, with the expert state-action tuples $\{(s_i, a_i)\}\in \tau_E^+$, our objective is to find the parameters $\phi_{init}$, which best fit the expert state-action pairs:

%
\begin{equation}
\label{equ: bc_training}
\phi_{init}=\argmax _{\phi} \prod \pi_{\phi}\left({a}_{i} | s_{i}\right)
\end{equation}
where $\pi_{\phi}({a}_{i} | s_{i})$ represent the probability of each expert action ${a}_{i}$ in the trained policy's Gaussian distribution $\pi_{\phi}(\cdot | s_{i})$. Therefore we can use Adam to solve $\phi_{init}$ based on the gradient to seek the maximum-likelihood. 

\subsection{Implementation}
The RL policy $\pi_\phi$ is realized by a deep neural network, which consists of a shared component of 6 CNN layers, MLP layers, and a separate policy head and value head of output size 6 and 1, respectively, as shown in Fig.~\ref{fig:overall framework}~(b).
Considering the balance of training speed and accuracy, the input size are scaled to $160\times128$ and the proximal policy optimization (PPO)~\cite{schulman2017proximal} is used.
Actions are scaled to [-1,1] by the tanh layer in the PPO policy network and then rescaled to their appropriate range [-1.5, 1.5] $mm$ and [-3$^{\circ}$, 3$^{\circ}$] to move the laparoscope.
The learning rate of PPO agent is 1e-5.


For discriminator $D_\theta$, the reward fuction $r_\theta$ is released by a basic reward network $r_{b}(\cdot)$ and shaping term $h(\cdot)$ with the same network structure but separate weights, consisting of 2 CNN layers followed by [32,32,1] MLP layers. 
The learning rate for the discriminator is 3e-4.
The batch size is 64, and the rollout buffer size is 4096.


\section{EXPERIMENT RESULTS}
\subsection{Experiment setup}
\begin{figure}[t]
  \centering
  \includegraphics[width = 1.0\hsize]{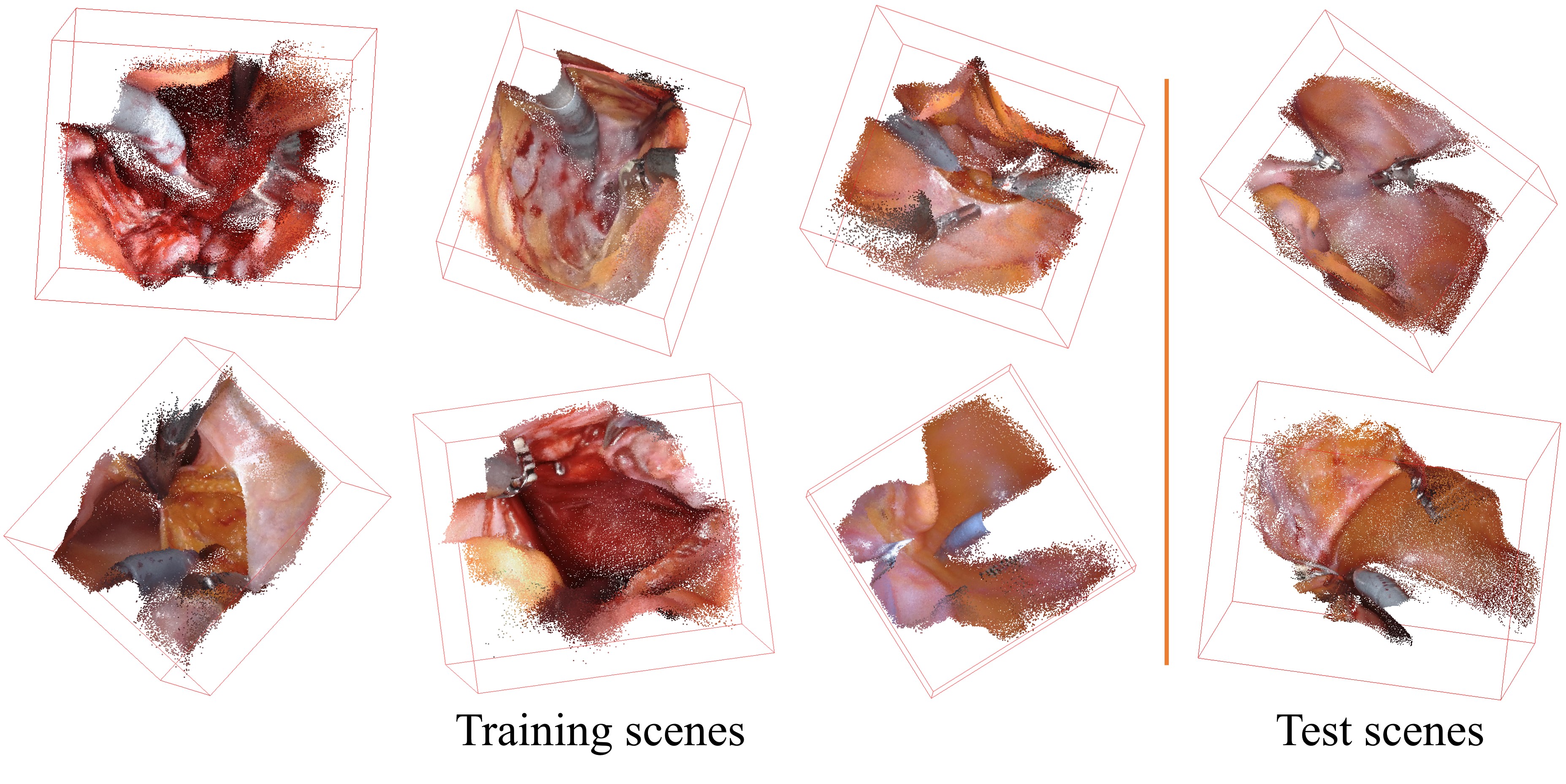}
  \caption{
  Screenshots of some typical training and test scenes.
  }
  \label{fig: envs}
\end{figure}

Since the 3D reconstruction may fail in severe conditions, such as lens contamination, bleeding, we manually check and collect 95 high-quality reconstructed surgical scenes as our dataset.
Among them, 80 scenes are randomly chosen for training and the rest 15 scenes for the test, as shown in Fig.~\ref{fig: envs}.
The statistics of video clips and reconstruction details (scene point cloud, trajectory) are shown in Table~\ref{table: env_statistics}.
On average, the laparoscope moves 8.23 $mm$ distance within 0.49 $s$, which shows the challenge of this task because it requires precise laparoscope control in a short time.

We built our proposed simulation environment based on the gym interface~\cite{1606.01540} and use \textit{PyTorch3D} as the render engine.
To improve the generality, the initial laparoscope position is randomly sampled in the origin trajectory workspace, also with the orientation variation.
An episode is terminated when the deviations of both the position and orientation from the expert endpoint are below particular thresholds, or when the episode steps reach a maximum length.
In our experiments, we set thresholds of position and orientation as 2 $mm$ and 5$^{\circ}$, and the maximum length to be 16.


\begin{table}[thpb]
\centering
\caption{Statistics of laparoscopy video clips and 3D reconstruction}
\label{table: env_statistics}
\begin{tabular}{c||ccccc} 
\hline
  Split & \begin{tabular}[c]{@{}c@{}}num\end{tabular} & 
  \begin{tabular}[c]{@{}c@{}}duration \\/ \textit{sec}\end{tabular} & 
  \begin{tabular}[c]{@{}c@{}}point number \\/ \textit{1e5}\end{tabular} & \begin{tabular}[c]{@{}c@{}}distance \\/ \textit{mm}\end{tabular} & steps  \\ 
\hline
Train & 80                                                           & 0.50±0.30                                                        & 1.92±0.53                                                          & 8.35±4.24                                               & 7.79±5.80     \\
Test   & 15                                                            & 0.42±0.11                                                       & 1.85±0.28                                                          & 7.57±4.22                                                & 6.33±3.17     \\ 
\hline
Total & 95                                                           & 0.49±0.28                                                        & 1.91±0.50                                                          & 8.23±4.25                                               & 7.56±5.50    \\
\hline
\end{tabular}
\vspace{-0.5cm}
\end{table}




\subsection{Evaluation metrics}
Success rate (SR) is used as the main metric to assess the performance of different policies.
%
%
Besides, we observe the motion pattern is not efficient in some successful episodes, e.g., the zig-zag movement.
%
Hence, we propose another metric action efficiency ($a_{eff}$) to measure the motion efficiency in successful episodes.
These metrics are calculated as:
%
%
%
\begin{equation}
\resizebox{0.9\hsize}{!}{$
\label{equ:success rate}
\begin{aligned}
\text{SR} = \frac{\sum_i^N sr_i}{N},\
sr = \mathbb{1}({\Delta p^T < \epsilon_p} \cap {\Delta r^T < \epsilon_r}), \\
\Delta p = \left\|P_{xyz} - P_{xyz}^{*}\right\|_2,\ \Delta r = cos^{-1}(\frac{tr(R_{\alpha \beta \gamma} \cdot R_{\alpha\beta\gamma}^{*})}{2})
\end{aligned}
$}
\end{equation}
%
%
\begin{equation}
\label{equ: act_eff}
a_{eff} = \frac {\sum_i^N \left({sr_i \cdot \left\|P_{xyz}^1 - P_{xyz}^{T_i}\right\|_2 / {T_i}} \right)}  {\sum_i^N {sr_i}}
\end{equation}
where $N$ is the episode number, $\mathbb{1}(\cdot)$ is the indicator function, $P^{(*)}$ and $R^{(*)}$ are the endpoint position and orientation in rollouts ($^{*}$ represent in demonstrations), $\epsilon_p$ and $\epsilon_r$ denote the corresponding thresholds, $tr(\cdot)$ is the trace of a matrix, $T$ is the total steps within one episode.

\begin{figure*}[t]
  \centering
  \includegraphics[width = 1.0\hsize]{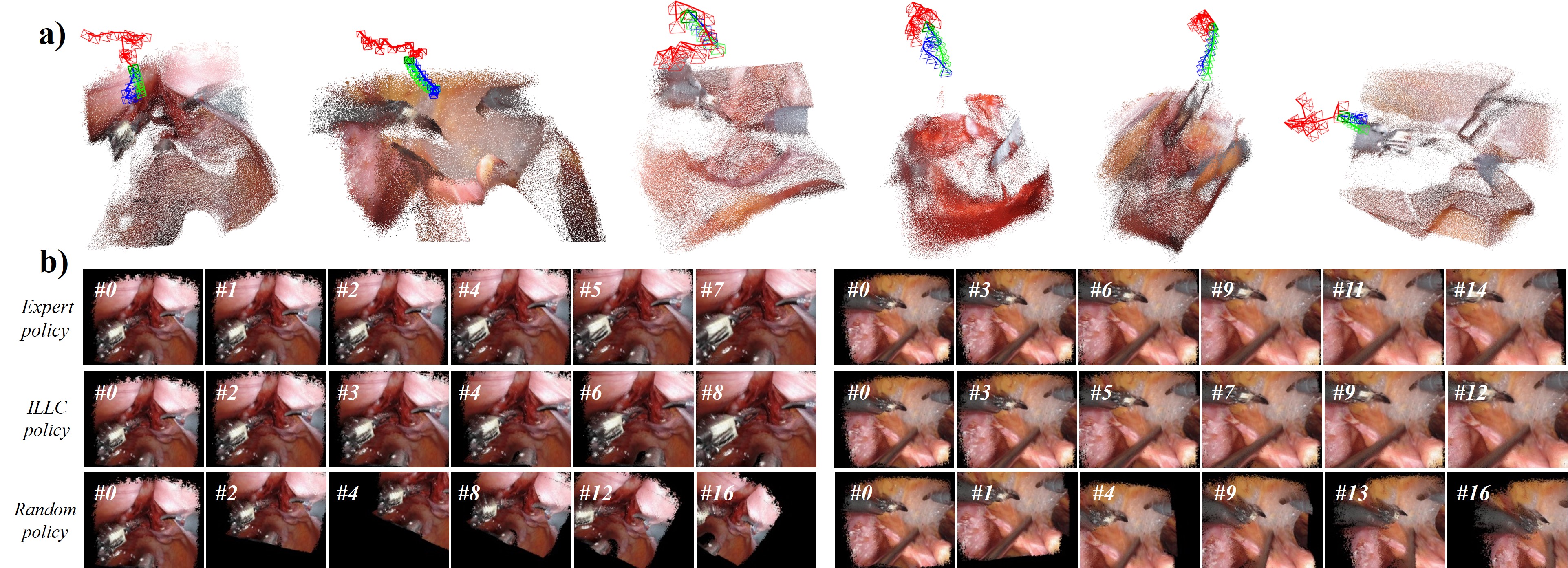}
  \caption{
  (a) Examples of laparoscope motion trajectories with different policies under test scenes: random, expert, trained policy are colored by red, green, blue respectively. (b) Screenshots of the \#\textit{steps} in the two test scenes running with expert (first row), ILLC (second row), random policies (last row). 
  }
  \label{fig: trajectory}
\vspace{-0.5cm}
\end{figure*}

\subsection{Results on validation}
%
\subsubsection{Comparison with IL baselines}
We run 50 episodes for policies with random initialization under each unseen test scene, and report the mean and the standard deviations for 15 test scenes.
The experimental results are listed in Table~\ref{table: validation}.
We choose the random policy with random actions to serve as a lower bound.
Two learning-based methods, i.e., behavior cloning (BC)~\cite{bojarski2016end} and GAIL~\cite{ho2016generative} are selected and trained with our augmented expert data for a fair comparison.
%
%
From the result, BC policy only achieves a success rate of 39.73\% with a large variance of 43.25\%, since it only remembers the actions in the training scenes and cannot generalize well to the test scenes.
GAIL policy has achieved a higher success rate of 43.86\%, however, it still fails in many cases as it is difficult to deal with unknown dynamics in various unseen scenes.
By contrast, our ILLC policy achieves the highest success rate of 66.47\% with the smallest variance and succeeds under a majority of unseen scenes, which demonstrates that the ILLC policy can learn the expert motion pattern with generality.
By illustrating the typical laparoscope trajectories generated by the random, expert, and our policy in Fig.~\ref{fig: trajectory}(a),
%
%
we can observe that our method can output similar laparoscope trajectories compared to the expert ones, and can also terminate with the same target view.
Moreover, even in some positions that deviated from the expert trajectories, our policy can still correct its direction and move to the corresponding ideal endpoint.
The screenshots in Fig.~\ref{fig: trajectory}(b) also indicate that the learned policy tend to achieve an appropriate FoV that is consistent with expert, e.g., making tools located in the middle region, and making the depth appropriate for the whole scene, etc.

\begin{table}[t]
\centering
\caption{Evaluation result of different policies on test scenes}
\begin{tabular}{c|ccc} 
\hline
                          Metrics & \begin{tabular}[c]{@{}c@{}}steps \\
                          (↓)\end{tabular} & \begin{tabular}[c]{@{}c@{}}$a_{eff}$ \\/ mm (↑)\end{tabular} & \begin{tabular}[c]{@{}c@{}}SR \\/ \% (↑)\end{tabular}  \\ 
\hline
Expert policy             & 5.63±3.45                                                 & 1.44±0.25                                                    & 100                                                              \\
\hline
Random policy             & 16.00±0.00                                                     & 0                                                            & 0                                                                \\
BC policy \cite{bojarski2016end}            & 10.98±6.06                                                & 0.90±0.78                                                    & 39.73±43.25                                                       \\
GAIL policy \cite{ho2016generative}            & 9.70±5.77                                                 & 0.86±0.53                                                     & 43.86±36.22                                                       \\
\hline
{\textbf{\textcolor{red}{ILLC (ours)}}} & \textbf{8.57±5.37}                                                 & \textbf{1.04±0.34}                                                 & \textbf{66.47±25.62}                                                     \\
\hline
\end{tabular}
\label{table: validation}
\vspace{-0.5cm}
\end{table}

\begin{table}[t]
\centering
\caption{Ablation study on policy training}
\label{table: ablation}
\begin{tabular}{c|ccc} 
\hline
                         Metrics & \begin{tabular}[c]{@{}c@{}}steps \\(↓)\end{tabular} & \begin{tabular}[c]{@{}c@{}}$a_{eff}$ \\~/ mm (↑)\end{tabular} & \begin{tabular}[c]{@{}c@{}}SR\\~/ \% (↑)\end{tabular}  \\ 
\hline
ILLC (SPTA@32x)            & 8.57±5.37                                                 & 1.04±0.34                                                    & 66.47±25.62                                                        \\ 

\hline
\textit{w/o} SPTA             & 10.26±5.32                                                  & 1.00±0.63                                                     & 48.27±38.38                                                       \\
\textit{w/} SPTA@8x             & 9.78±4.70                                                  & 1.23±0.44                                                    & 54.00±31.73                                                      \\
\hline
\textit{w/o} prior policy & 10.57±5.37                                                 & 0.89±0.32                                                     & 46.21±39.20                                                    \\ 
\hline
\textit{w/o} depth       & 9.83±5.12                                                  & 0.98±0.56                                                     & 54.29±39.34                                                        \\
\hline
\end{tabular}
\vspace{-0.4cm}
\end{table}

\begin{figure}[t]
  \centering
  \includegraphics[width = 1.0\hsize]{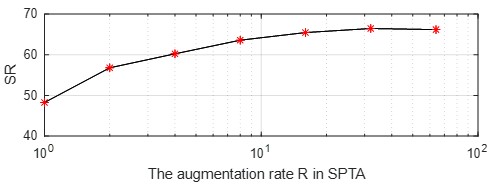}
  \caption{
  Policy performance with SPTA under different augmentation rates.
  }
  \label{fig: spta}
\vspace{-0.5cm}
\end{figure}

\subsubsection{Ablation study}
We design experiments to analyze the impact of three factors: 1) SPTA, 2) prior policy, 3) observed state.
%
The policy performance with SPTA under different augmentation rates $R$ (SPTA@$R$x) is compared in Table~\ref{table: ablation} and also visualized in Fig.~\ref{fig: spta}.
%
%
Results show that SPTA can largely improve the success rate, e.g., from 48.27\% without SPTA to 66.47\% with SPTA@32x, of a smaller variance and improved action efficiency.
The reason is that SPTA can generate more expert-like data, which enables the IL agent to fully explore the environment and achieve a better generality.
%
Table~\ref{table: ablation} also compares the policy trained without the prior policy,
we can observe that the prior policy can improve the success rate by 20.26\% since it can provide more meaningful rollouts in the early stage and bootstraps the overall training process, hence proving its importance in our policy.
%
Finally, we discuss the impact of depth in the state representation on the learned policy.
Results show that including the depth in the input can improve the success rate by 12.18\%, which demonstrates its benefits to the FoV control.


\subsubsection{Visualization analysis}
\begin{figure}[t]
  \centering
  \includegraphics[width = 1.0\hsize]{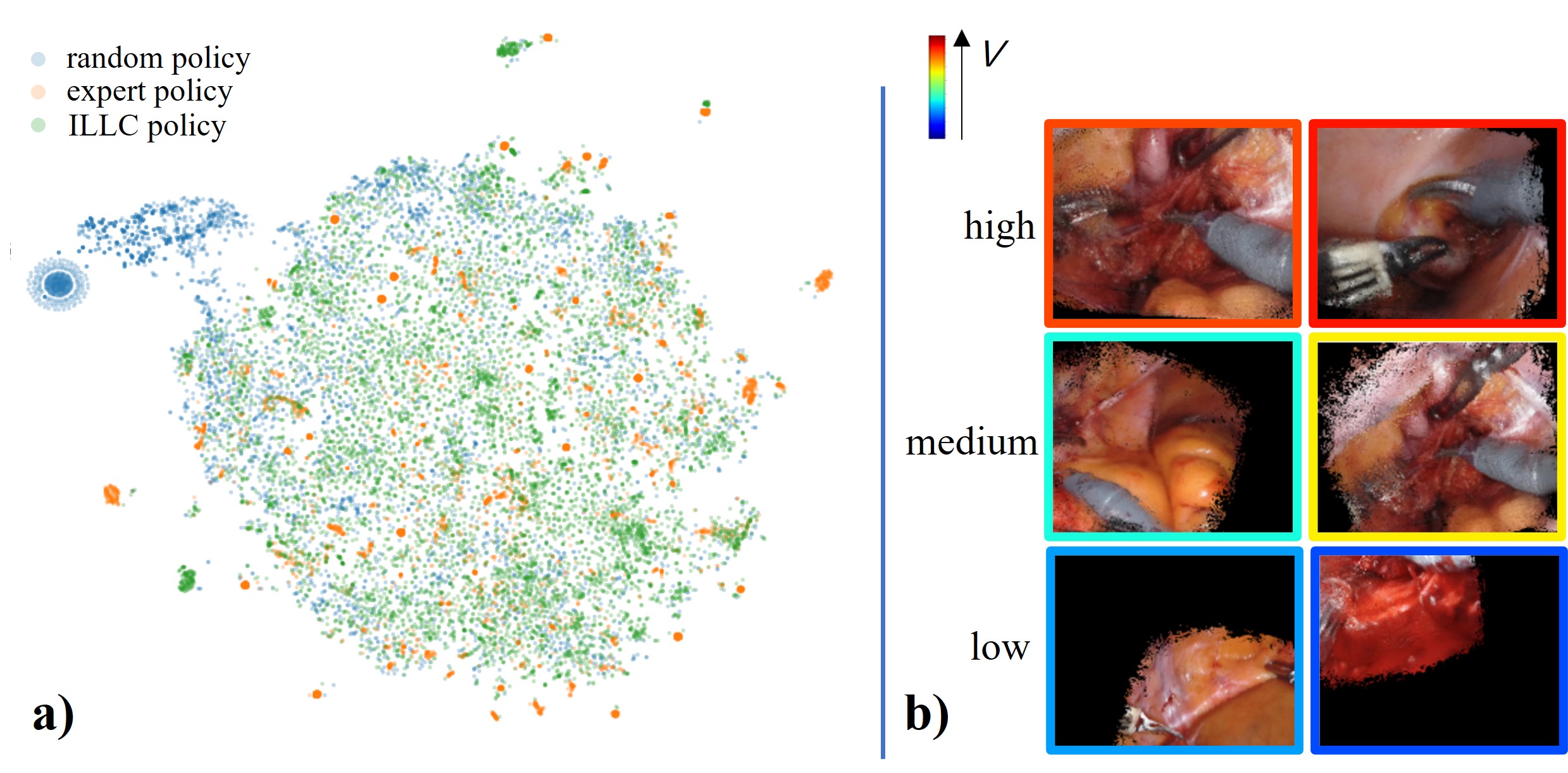}
  \caption{(a) 2D t-SNE embeddings of policy network features for data collected by different policies: random, expert, and our ILLC policy.
  (b) Visualization of predicted values by trained policy; the color surrounding the image represents the state values from lowest (blue) to highest (red).
  }
  \label{fig: t-SNE}
\vspace{-0.5cm}
\end{figure}

To explore the hidden knowledge learned by our proposed ILLC policy when performing the laparoscope control task and intuitively illustrate such information,
we use t-SNE~\cite{van2008visualizing} to visualize the embedding features in the last hidden layer of the ILLC policy network.
The input observed states to the network are collected in test scenes for evaluation with random, expert, and ILLC policy, respectively.
As shown in Fig.~\ref{fig: t-SNE}(a), the distributions of ILLC policy and expert policy are thoroughly mixed and far away from the random policy ones,
which indicates that to same extent the ILLC policy controls the laparoscope in a similar way to the experts.
%
%
We also visualize the estimated values from the policy network with observed images in Fig.~\ref{fig: t-SNE}(b), where high values indicate the observed states are promising outcomes.
As expected, the ILLC policy maps the proper laparoscopic views to high estimated values across different test scenes.
For example, in Fig.~\ref{fig: t-SNE}(b), the top row with appropriate views has higher state values, and the bottom row far from the ideal viewpoints has low values.
%

\section{CONCLUSIONS}
In this paper, we propose a novel ILLC framework to achieve intelligent autonomous laparoscope control.
Unlike conventional tracking-based control approaches, the learned policy generates actions considering the RGB-D surgical scene context. 
Meanwhile, the policy can capture the joint connection between the laparoscope motion and observed scenes from raw surgical videos.
The learned policy shows superior laparoscope control performance and generality to unseen test scenes compared with previous IL baselines.

In the future, we will enrich the surgical scenes from surgical videos and improve the 3D reconstruction quality to make the interactive environment more realistic.
In addition, we will optimize the policy network, such as adding temporal information to further improve performance.

\bibliographystyle{IEEEtran}
\bibliography{ref}

\end{document}